\documentclass[conference]{IEEEtran}
\IEEEoverridecommandlockouts

\usepackage{tikz}
\usepackage{textcomp}
\usepackage{hyperref}
\usepackage{lipsum}

\usepackage[T1]{fontenc}
\usepackage[utf8]{inputenc}
\usepackage{authblk}

\usepackage[backend=bibtex, style=numeric, sorting=none]{biblatex}

\title{More than one Author with different Affiliations}
\author[1]{Hafiq Anas}
\author[2]{Bacha Rehman}
\author[1]{Wee Hong Ong}
\affil[1]{Robotics and Intelligent Systems Lab, Faculty of Science, Universiti Brunei Darussalam, Brunei}
\affil[2]{Department of Computer Science, Namal Institute, Mianwali, Pakistan}

\begin{document}
%
\title{Deep Convolutional Neural Network Based Facial Expression Recognition in the Wild}
\maketitle

\begin{abstract}
This paper describes the proposed methodology, data used and the results of our participation in the Challenge-Track 2 (Expr Challenge Track) of the Affective Behavior Analysis in-the-wild (ABAW) Competition 2020. In this competition, we have used a proposed deep convolutional neural network (CNN) model to perform automatic facial expression recognition (AFER) on the given dataset. Our proposed model has achieved an accuracy of 50.77\% and an F1 score of 29.16\% on the validation set. 
\end{abstract}

\section{Introduction}
Automatic facial expression recognition (AFER) has been a challenging task given the inherent problems in computer vision as well as the significant ambiguity in the expression of different subjects with different intensity.  Different methods \cite{4468714}\cite{article} have been proposed to develop AFER systems. These methods include featureless approaches such as the Convolutional Neural Network (CNN) that directly work on the images, and feature-based approaches such as the Deep Neural Network (DNN), Support Vector Machine (SVM), Local Binary Patterns (LBP), and Facial Action Units (AU).

CNN models can usually be architectured and tuned to fit a good model on a given dataset without feature engineering. This has led to the popular use of the CNN models in various computer vision tasks including facial expression recognition. In this paper, we have used a CNN-based AFER model that only takes raw image pixel information input for expression classification. The proposed model has been used to make predictions for Challenge-Track 2 (Expr Challenge-Track) in the Affective Behavior Analysis In-The-Wild (ABAW) competition 2020 \cite{kollias2020analysing}. Track 2 requires the proposed model to make an accurate classification of the seven basic expressions from images. The seven basic expressions are neutral, anger, disgust, fear, happiness, sadness, and surprise.

Affective Behavior Analysis In-The-Wild (ABAW) 2020 is a competition that aims to advance the current state-of-the-art problem in the analysis of human effective behavior in-the-wild. For the first time, the competition contains three challenges which are: valence-arousal estimation, seven basic expression classification, and facial action unit detection. This competition provides the largest wild dataset which contains images, videos, and audio data along with three types of annotated labels according to the three challenges. The seven basic facial expressions are one of the most frequently used emotion representations for Facial Expression Recognition (FER) research and our proposed method uses the seven basic facial expression labels for expression classification.

\section{Proposed Method}
In this competition, we have used a publicly available CNN-based deep learning model from GitHub user named Mayur Madnani \cite{mayurmadnani2018} that was developed for the FER2013 dataset \cite{carrier2013fer} during the Kaggle’s “Challenges in Representation Learning: Facial Expression Recognition Challenge” in 2013 \cite{fer2013challenge_report}. The model achieved a testing accuracy of 65.5\% which was comparable to getting the fifth position in the challenge. FER2013 is a large in-the-wild dataset with 35,887 images that was introduced during the International Conference of Machine Learning conference in 2013 (ICML 2013) and used in Kaggle’s “Challenges in Representation Learning: Facial Expression Recognition Challenge” in 2013. This dataset is divided into two sets: the training set (28,709 images) and the testing set (3,589 images). Each image was manually annotated with the seven facial expression labels.

\subsection{Network Architecture}
\begin{figure}[!htb]
	\centering
	\includegraphics[width=1.0\linewidth]{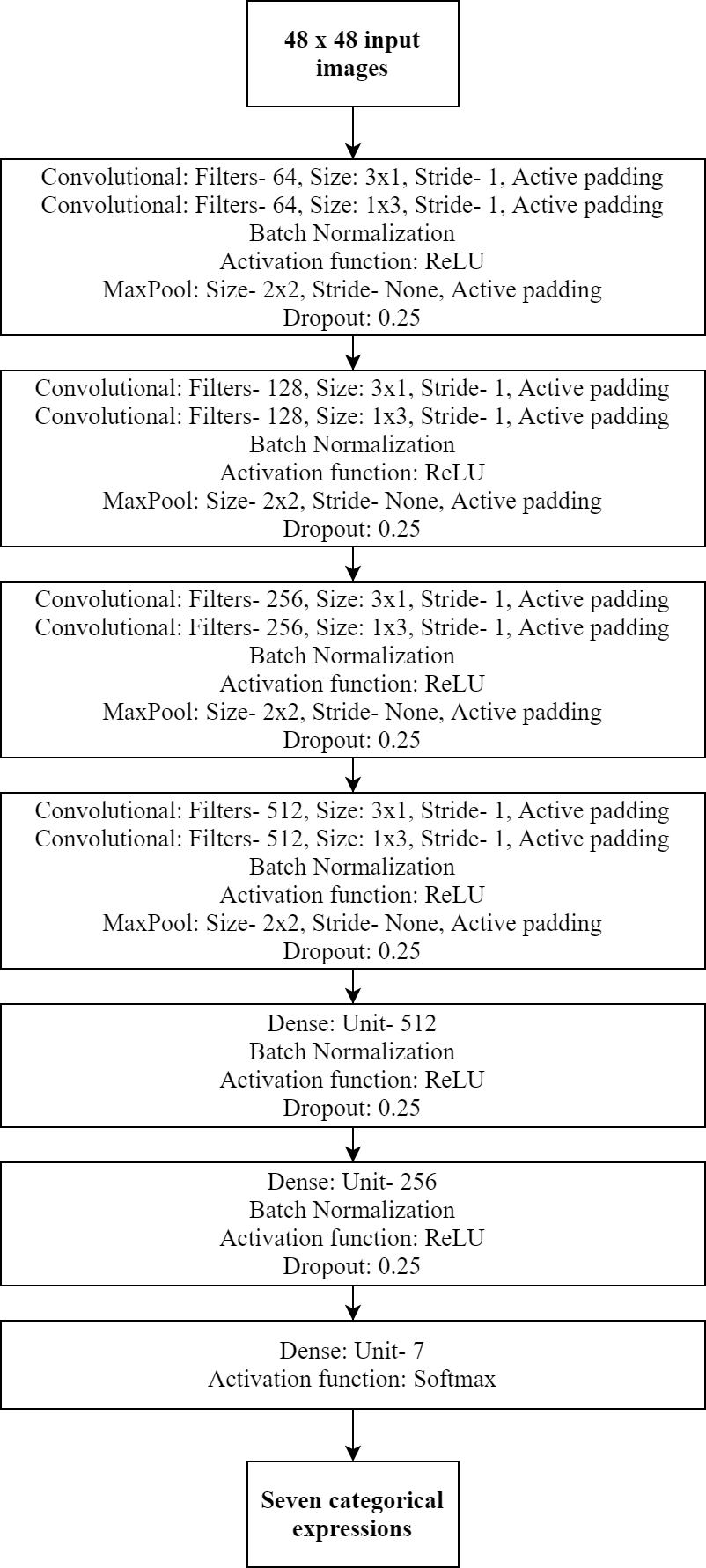}
	\caption{Mayur Madnani's Deep CNN architecture}
	\label{fig:fig1}
\end{figure}

The network architecture consists of seven phases as shown in \autoref{fig:fig1}. Each of the first four phases contains two Convolutional2D layers with a filter size of 64, 128, 256, and 512 respectively and uses single MaxPooling2D layers with a pool size of 2 by 2, no strides, and no padding. For the two Convolution2D layers, a kernel size of 3 x 1 and 1 x 3 were used in order with no padding.
The remaining three phases contain a Dense layer of 512, 256, and 7 units in which the final unit corresponds to the seven facial expression predictions. A Flatten layer is added in between the first four phases and the last three phases.
Except for the last phase, a BatchNormalization layer and an activation function of ReLu were added. The last phase used Softmax as the activation function. The model is then compiled using Adam as the optimizer.

\section{Experiment Details}
This section contains Aff-Wild2 dataset information and our experiment procedures. The entire training and testing is done on a PC with the following specifications: Intel i7 9700 CPU, RTX 2070 GPU and 64GB DDR4 RAM.

\subsection{Aff-Wild2 dataset}
Aff-Wild2 is by far the largest available wild dataset consisting of 545 videos with 2,786,201 frames. It is an extension to the previous version which was the Aff-Wild dataset \cite{kollias2018aff}\cite{kollias2019expression} \cite{kollias2019deep}\cite{zafeiriou2017aff}. This dataset provides cropped and aligned facial images that are useful for participants with no access to a face detector. Additionally, the dataset contains annotations in the form of seven basic facial expressions, facial action units, and valence-arousal intensity. The organizers of the competition have given training, validation, and testing sets for all tracks.

\subsection{Data preparation and preprocessing}
In our experiment approach, we have merged both the cropped and aligned images of training and validation sets and did a train: test split ratio of 80:20. The Aff-Wild2 dataset was split into two sets: training set (80\% of the merged dataset) and validation set (remaining 20\% of the merged dataset). The dataset is then preprocessed by converting each image to greyscale, resizing to 48 x 48 pixels, and then converted to NumPy format for baseline model training and testing. 

\subsection{Training setup and results}
The CNN model is then trained on the training set with 100 epochs, a batch size of 512, a learning rate of 0.001, and a weight decay of 0.000001. Finally, the model is tested on the validation set. An accuracy of 50.77\% with an F1-score of 29.16\% was achieved on the validation set.

\section{Conclusion}
In this paper, we have used a CNN-based deep learning model that was fine-tuned using FER2013 dataset. Currently, Aff-Wild2 is the largest and latest dataset which provides the necessary data (i.e image, video, audio) for Facial Expression Recognition (FER) based research on valence-arousal estimation, facial action unit detection, and seven basic expression classification. The FER2013 model has achieved an accuracy of 50.77\% on the Aff-Wild2 test set with image pixels information alone, which leaves big room for improvement. A potential improvement is the addition of DNN (Deep Neural Network) that takes in facial landmarks information and can be concatenated to the CNN model to form a hybrid CNN-DNN model for a more robust facial expression classification.

\section{Acknowledgment}
We gratefully acknowledge Mayur Madnani for making their CNN model publicly available on their GitHub page. We are thankful for the support from Dimitrios Kollias in promptly responding to our queries.

\printbibliography

\end{document}